\documentclass[10pt,journal,compsoc]{IEEEtran}

\ifCLASSOPTIONcompsoc
  \usepackage[nocompress]{cite}
\else
  \usepackage{cite}
\fi

\usepackage{times}
\usepackage{epsfig}
\usepackage{graphicx}
\usepackage{amsmath}
\usepackage{amssymb}

\usepackage{comment}
\usepackage{algorithmic}
\usepackage{graphicx}
\usepackage{textcomp}
\usepackage{epsfig}
\usepackage{subcaption}

\usepackage{rotating}

\usepackage{enumitem}

\def\BibTeX{{\rm B\kern-.05em{\sc i\kern-.025em b}\kern-.08em
    T\kern-.1667em\lower.7ex\hbox{E}\kern-.125emX}}
    
\newcommand{\matr}[1]{\mathbf{#1}}
\DeclareMathOperator*{\minimize}{minimize}

\newcommand{\etal}{\emph{et al.}}

\usepackage{multirow}
\usepackage{tikz}
\usetikzlibrary{positioning}
\usetikzlibrary{calc}
\usetikzlibrary{fit}
\usepackage{dblfloatfix}

\usepackage{mathtools}

\usepackage[pagebackref=true,breaklinks=true,colorlinks,bookmarks=false]{hyperref}

\begin{document}

\title{Canonical Face Embeddings}

\ifCLASSOPTIONpeerreview
    \author{Blind
    
    \thanks{Manuscript received XXXXXXXX.}}

    \markboth{IEEE Transactions on Biometrics, Behavior, and Identity Science, Vol. X, No. X, April 2021}%
    {Blind Submission: Canonical Face Embeddings}
\else
    \author{David~McNeely-White,
            Ben~Sattelberg,
            Nathaniel~Blanchard,
            and~Ross~Beveridge,
    \IEEEcompsocitemizethanks{\IEEEcompsocthanksitem All authors are with the Department
    of Computer Science, Colorado State University, Fort Collins, Colorado.\protect\\
    E-mail: \{david.white, ben.sattelberg, nathaniel.blanchard, ross.beveridge\}@colostate.edu
    \IEEEcompsocthanksitem Code to reproduce these findings will be released upon publication.
    }
    \thanks{Manuscript received XXXXXXXX.}}

    \markboth{IEEE Transactions on Biometrics, Behavior, and Identity Science, Vol. X, No. X, April 2021}%
    {McNeely-White~\etal: Canonical Face Embeddings}
\fi

\IEEEtitleabstractindextext{%
\begin{abstract}
We present evidence that many common convolutional neural networks (CNNs) trained for face verification learn functions that are nearly equivalent under rotation.
More specifically, we demonstrate that one face verification model's embeddings (i.e. last--layer activations) can be compared directly to another model's embeddings after only a rotation or linear transformation, with little performance penalty.
This finding is demonstrated using IJB-C 1:1 verification across the combinations of ten modern off-the-shelf CNN-based face verification models which vary in training dataset, CNN architecture, method of angular loss calculation, or some combination of the 3.  These networks achieve a mean true accept rate of 0.96 at a false accept rate of 0.01.
When instead evaluating embeddings generated from two CNNs, where one CNN's embeddings are mapped with a linear transformation, the mean true accept rate drops to 0.95 using the same verification paradigm.
Restricting these linear maps to only perform rotation produces a mean true accept rate of 0.91.
These mappings' existence suggests that a common representation is learned by models despite variation in training or structure.
We discuss the broad implications a result like this has, including an example regarding face template security.
\end{abstract}

\begin{IEEEkeywords}
Representational similarity, de-anonymization, facial recognition, feature space mapping, neural network equivalence
\end{IEEEkeywords}}

\maketitle

\IEEEdisplaynontitleabstractindextext

\IEEEpeerreviewmaketitle

\IEEEraisesectionheading{\section{Introduction}\label{sec:introduction}}

\IEEEPARstart{T}{he} last decade of research into neural networks can be coarsely categorized into efforts toward: 1) advancing the state-of-the-art as expressed through accuracy, and 2) better understanding and analyzing what networks learn.
Efforts toward understanding have been far ranging: from comparisons to human vision \cite{blanchard2019neurobiological,storrs2020diverse}, to investigations into the quality and content of learned features \cite{hermann2020origins}, to detailed breakdowns of what causes networks to fail \cite{dodge2016understanding}. Even research areas as distinct as style transfer~\cite{hart2020style,gatys2016image} and transfer learning \cite{oquab2014learning} are all, from a certain vantage, centered around understanding the nature of what CNNs learn and, in turn, how what they learn separates and semantically organizes information. 
Indeed, such fundamental understanding is crucial in the pursuit of better performing and more predictable machine learning models. 

\begin{figure}[t]
  \centering
  \includegraphics[width=0.45\textwidth]{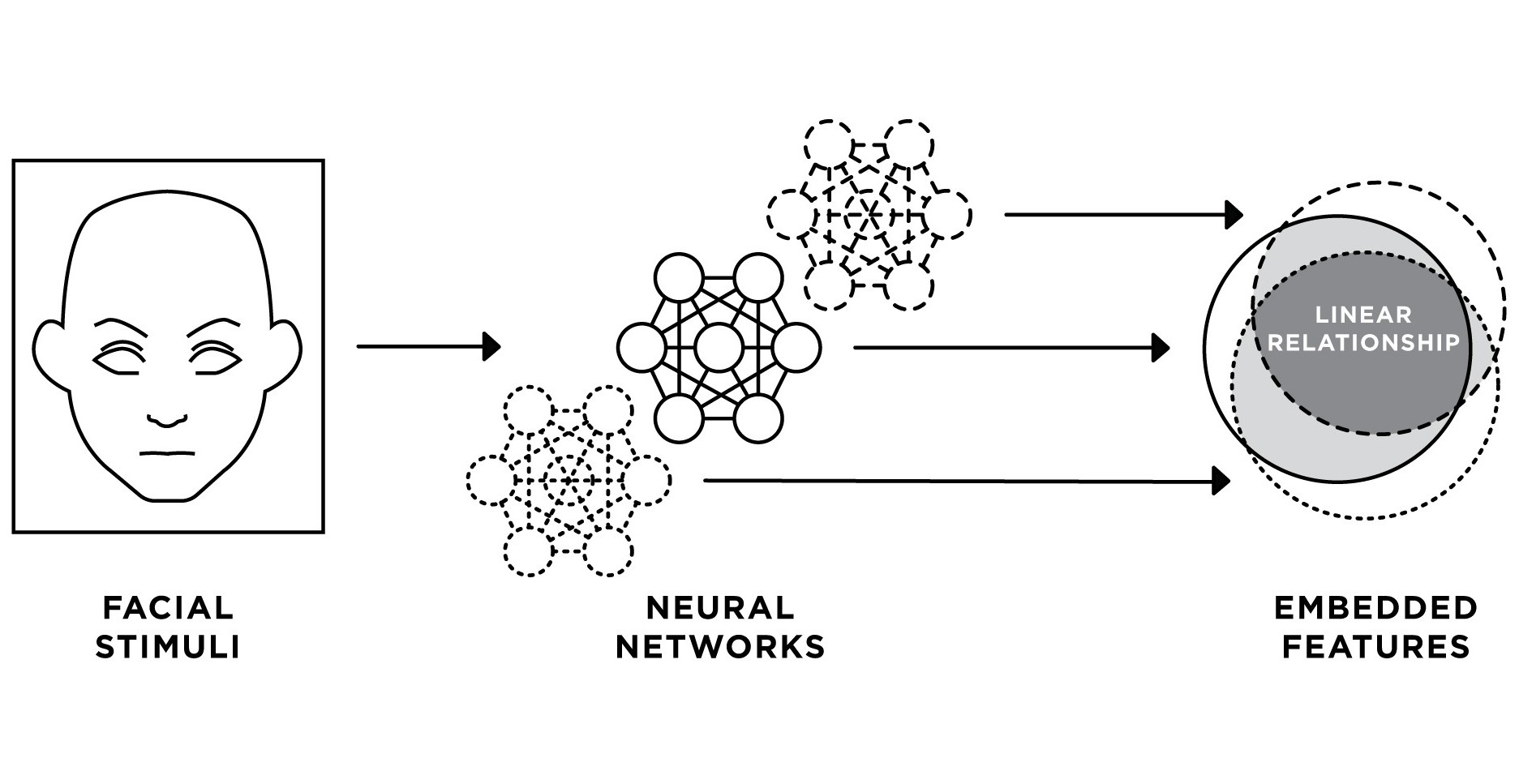}
  \caption{CNN-based facial recognition systems produce embedding spaces which are typically used for distance-based face verification.  The embedding spaces generated by different networks are out-of-the-box not directly comparable. However, our work demonstrates the existence of an underlying common geometry such that we are able to compute a linear mapping to recover the embeddings of one network from another. This allows direct meaningful comparison between pairs of embeddings originating from different networks.}
  \label{fig:teaser}
\end{figure}

Our work here focuses specifically on the embeddings generated by different CNNs. To avoid confusion based on a number of conflicting definitions for neural network ``embeddings,'' we use the term embedding to refer to the feature vector generated by the pooled output of the final, typically convolutional, layer. In closed-set classification tasks, these embeddings are typically fed to a final linear and fully-connected classification layer with one output unit for each class label in the dataset. In open-set face recognition, these embeddings are typically used in a distance-based nearest-neighbors approach instead.

Face recognition is an excellent domain for
measuring if embeddings from different networks are equivalent, i.e. related through a simple linear mapping, in part because it is a domain with strongly established protocols, available data, and an emphasis on direct comparison of embeddings.  In particular, the open-set verification protocol (i.e. no people in common between training and testing) removes the risk that measured equivalence is a consequence of a common final logit layer with fixed labels shared between all networks. Indeed, our prior work \cite{mcneely2020same,mcneely2020exploring} has established that networks trained on matching label sets such as ImageNet have linear mappings between embedding spaces which may be computed directly from the final layer weights of the two networks. Going further, the now-standard use of unit-normalized features compared by cosine distance ~\cite{arcface, wang2018additive, wang2017normface, wang2018cosface} provides a clear sense in which network outputs are the same or not.
In essence, this allows us to ask questions of the kind `Is this geometric object A the same as object B?''.  For rigid objects, if B is just A rotated, the common answer is they are the same. The work here takes this basic concept of ``same'' (or equivalent) and applies it to face embeddings from two CNNs. 

Using as few as a hundred embedding pairs generated from two nets for the same faces, our approach computes mappings using least-squares regression in order to align, as much as is possible, the corresponding embedding vectors.
These mappings are evaluated by comparing the embeddings of one CNN to the mapped embeddings of another using a standard face verification paradigm.
Despite differences in training dataset, CNN architecture, and angular loss function, cross-CNN face verification using these mappings drops very little. 
These results demonstrate a fundamental underlying near equivalence between the embedding spaces produced by diverse models.
We describe this near equivalence through the lens of a canonical embedding space,  
i.e. that different models' embedding sets are sampled from linear transformations of a single underlying embedding space.
As we discuss in Section~\ref{sec:conclusion}, the existence of this canonical embedding space suggests that task is the primary driver in determining the content of learned representations.

We want to emphasize that we essentially never see improved verification results using features mapped between networks, nor do we expect such improvement.  Our goal is not to build a better algorithm.  Instead, when interpreting the results which follow and which take advantage of established standard evaluation protocols, the point is to demonstrate the relatively small drop in recognition performance when mapping between spaces and use this observation as very strong evidence for the existence of a common underlying embedding space geometry. 

Our work complements a growing sense that different architectures applied to the same data are converging upon similar solutions. For example, some within the neural architecture search community have shifted their focus from improving the architecture itself to finding new data augmentation strategies \cite{cubuk2019autoaugment, zoph2019learning, cubuk2020randaugment}. Our discovery of an underlying canonical embedding space also strongly suggests that recovery of identities from embeddings is possible, and even arguably easy in some cases: for example using embeddings from the specific CNNs studied here.  

\section{Related Work}
\label{sec:related}
Many have studied neural networks for the purpose of understanding their hidden representations.
Techniques range from visualization \cite{erhan2009visualizing, zeiler2014visualizing, olah2017feature}, to semantic interpretation \cite{kim2017interpretability, zhou2016learning}, to similarity metrics. 
Efforts using similarity metrics are most aligned with our goals.

Li~\etal\cite{li2015convergent} used matching algorithms to align individual units of ImageNet-trained deep CNNs to determine if networks of different initializations converge to similar representation.
This work is one of the earliest efforts along the same lines as ours.
They find evidence that ``some features are learned repeatedly in multiple networks, but other rare features are not always learned.''
However, drawing semantic meaning from individual bases may be problematic --- others have suggested that semantic meaning is contained within the entire feature space, rather than individual units \cite{szegedy2013intriguing}.
Wang~\etal\cite{wang2018towards} expands on this effort by searching for minimal subsets of neurons which linearly represent one another. 
Surprisingly, they report near-zero similarity for all layer-wise comparisons of the models they study, in contrast with both later representational similarity efforts (covered next) and our work.
This may be due to their comparison of spatially-sensitive convolutional layer activations, or their restriction to using subsets of neurons instead of complete representations.

Correlation-based methods have also been used for studying representational similarity, such as canonical correlation analysis (CCA), singular vector CCA (SVCCA), projection weighted CCA, or centered kernel alignment \cite{raghu2017svcca, morcos2018insights, kornblith2019similarity}. 
These studies have revealed a number of interesting deep neural network properties, including the tendency of layers to be learned in a bottom-up fashion \cite{raghu2017svcca}, a correspondence between networks' size and relative similarity \cite{morcos2018insights}, and similarity both between and within networks trained on different datasets \cite{kornblith2019similarity}. 
The latter work is particularly relevant, as it focuses on the degree of representational similarity between networks which differ in parameterization, much as our own does. 
While correlation-based techniques may be invariant to orthogonal transform, linear transform, or isotropic scaling, and thus revealing of types of relationships between feature representations, the metric they provide is ultimately subjective.
Kornblith~\etal~\cite{kornblith2019similarity}  address this by evaluating similarity indexes based on their ability to accurately identify whether two CNN representations belong to the same layer of architecturally identical networks trained from different random initialization.
In contrast, our work provides a \textit{parameterization} of the relationship between feature representations (i.e., a mapping matrix), which can be evaluated using the same methods used to evaluate the models in question.
Consequently, we can compare the performance of features which have been mapped against those which have not for a clearer and more objective baseline.

To our knowledge, only two other works have used linear transformations to compare hidden deep neural network representations.
Lenc and Vedaldi \cite{Lenc2019} previously investigated affine mappings between layers of AlexNet \cite{krizhevsky2012imagenet}, VGG-16 \cite{simonyan2013deep}, and ResNet-v1-50 \cite{he2016deep} on ILSVRC2012.
However, they fit mappings between spatially-sensitive convolutional layers, and note ``there is a correlation between the layers’ resolution and their compatibility.'' 
Further, they fit mappings using the same classification loss used during training of the models they study, amounting to a more coarse-grained optimization target, and resembling the fixed-feature extraction used for transfer learning.
Still, these efforts provide an exciting foundation for us to build upon.
In our work, we map between the spatially-pooled outputs of the final convolutional layer of each model, requiring no lossy spatial interpolation, and use pairs of embeddings and a distance-based loss to fit maps.

Our own prior work, 
McNeely-White~\etal\cite{mcneely2019inception}, demonstrated linear similarity between feature representations of two ILSVRC2012-trained CNNs despite differences in architecture (Inception-v4 and ResNet-v2 152).
We have since expanded this finding to cover 10 different ILSVRC2012 models \cite{mcneely2020same, mcneely2020exploring}.

While clearly encouraging to us, efforts such as these (along with many of those mentioned earlier in this section) may be problematic. 
Namely, there is a degree to which similarity should be expected between embedding spaces for models trained on not only the same task, but the same \textit{labels}.
In essence, as long as a linear classifier is used to convert hidden representations into same set of labels, cross-CNN similarity is explicitly encouraged during training by softmax cross-entropy loss.
This topic is described in greater detail in \cite{mcneely2020exploring}.
Further, Roeder~\etal \cite{roeder2020linear} recently established a theoretical proof for the existence of linear correspondence between deep supervised classification models (among others).

Consequently, face verification is an excellent task domain for facilitating linear correspondence studies such as this.
Even though models may use a linear classifier with softmax cross-entropy loss during training, we include models which do not have overlapping training labels.
What's more, we do not map between faces in the training set, instead mappings are computed using a testing dataset (IJB-C) which is disjoint from the data used to train networks.
While linear correspondence between closed-set image classifiers is certainly compelling, by establishing linear correspondence between open-set image classifiers, we establish even stronger evidence that a fundamental similarity in learned representations exists between modern deep CNNs. 

\section{Methods}
\label{sec:method}
\begin{table}
  \centering
  \renewcommand{\arraystretch}{1.1}
  \begin{tabular}{l|l|l}
  Dataset & \# individuals & \# images/video frames \\\hline
  VGGFace2 \cite{vggface2} & 9.1 K & 3.3 M \\
  CASIA-WebFace \cite{CASIAwebface} & 10.5 K & 0.5 M \\
  MS1M \cite{ms-celeb-1m} & 100 K & 10 M \\
  MS1MV2\footnote{Refined version of MS1M.} \cite{arcface} & 85 K & 5.8 M \\  
  Glint360K \cite{an2020partial_fc} & 360 K & 17 M \\
  \textbf{IJB-C} \cite{maze2018iarpa} & \textbf{3.5 K} & \textbf{148.8 K} 
  \end{tabular}
  \caption{The datasets used in our experiments. The first five datasets were used to train networks used in our experiments. The final dataset, \textbf{IJB-C}, is a test dataset used to test mappings between feature spaces. }
  \label{tab:dataset-desc}
\end{table}

\begin{table}
    \centering
    \renewcommand{\arraystretch}{1.1}
    \begin{tabular}{l|l|l}
        Model & \# parameters & Flops \\\hline
        MobileNet-v2 1.4 224 \cite{sandler2018mobilenetv2} & 6.1 M & 1.2 B \\
        64-CNN \cite{liu2017sphereface} & 23 M & 3.5 B \\
        Inception-ResNet-v1 \cite{szegedy2017inception} & 24 M & 11 B \\
        ResNet-50 \cite{he2016identity} & 25.6 M & 7.2 B \\
        ResNet-100 \cite{he2016identity} & 44.5 M & 13.4 B \\
    \end{tabular}
    \caption{Computational size and complexity of CNN backbones studied. 
    }
    \label{tab:backbones}
\end{table}

\begin{table*}
  \centering
  \renewcommand{\arraystretch}{1.3}
  \setlength\tabcolsep{4pt}
    \begin{tabular}{|l|l|l|l|c|c|c|c|c|c|l|}
    \hline
        \multicolumn{4}{|c}{} & \multicolumn{6}{|c|}{\textbf{IJB-C (TAR @ FAR)}} & \\
        \hline
        \textbf{Short Name} & \textbf{Training Dataset} & \textbf{CNN Architecture} & \textbf{Angular Loss Function} & \textbf{1e-1} & \textbf{1e-2} & \textbf{1e-3} & \textbf{1e-4} & \textbf{1e-5} & \textbf{1e-6} & \textbf{Source} \\
    \hline
    M2\_R100\_A & MS1MV2 & ResNet100 \cite{he2016identity} & ArcFace \cite{arcface}& 0.991 & 0.984 & 0.975 & 0.963 & 0.945 & 0.898 & \cite{insightfaceGithub} \\
    \hline
    V\_R50\_A & VGGFace2 & ResNet50 \cite{he2016identity} & ArcFace \cite{arcface}& 0.994 & 0.984 & 0.963 & 0.928 & 0.875 & 0.744 & \cite{insightfaceGithub} \\
    \hline
    G\_R100\_P1.0 & Glint360k & ResNet100 \cite{he2016identity} & PartialFC (r=1.0) \cite{an2020partial_fc} & 0.993 & 0.988 & 0.981 & 0.973 & 0.960 & 0.912 & \cite{insightfaceGithub} \\
    \hline
    G\_R100\_P0.1 & Glint360k & ResNet100 \cite{he2016identity} & PartialFC (r=0.1) \cite{an2020partial_fc} & 0.992 & 0.987 & 0.981 & 0.974 & 0.961 & 0.872 & \cite{insightfaceGithub} \\
    \hline
    M1\_R50\_A & MS1M & ResNet50 \cite{he2016identity} & ArcFace \cite{arcface} & 0.979 & 0.954 & 0.918 & 0.861 & 0.782 & 0.701 & \cite{arcfaceGithub}* \\
    \hline
    M1\_MB2\_A & MS1M  & MobileNetV2 \cite{sandler2018mobilenetv2} & ArcFace \cite{arcface} & 0.981 & 0.940 & 0.869 & 0.766 & 0.629 & 0.503 &  \cite{arcfaceGithub}* \\
    \hline
    V\_IR1\_C & VGGFace2 & InceptionResNetV1 \cite{szegedy2017inception} & Center Loss \cite{centerLoss}& 0.990 & 0.967 & 0.908 & 0.808 & 0.681 & 0.518 &  \cite{facenetGithub}* \\
    \hline
    C\_IR1\_C & CASIA-WebFace & InceptionResNetV1 \cite{szegedy2017inception} & Center Loss \cite{centerLoss} & 0.981 & 0.929 & 0.832 & 0.697 & 0.534 & 0.408 &  \cite{facenetGithub}* \\
    \hline
    M1\_64S\_PFE & MS1M & 64-CNN+PFE \cite{liu2017sphereface, shi2019probabilistic} & AM-Softmax \cite{wang2018additive} & 0.985 & 0.970 & 0.942 & 0.872 & 0.757 & 0.610 &  \cite{pfeGithub} \\
    \hline
    C\_64S\_PFE & CASIA-WebFace & 64-CNN+PFE \cite{liu2017sphereface, shi2019probabilistic} & AM-Softmax \cite{wang2018additive} & 0.982 & 0.949 & 0.889 & 0.798 & 0.678 & 0.530 & \cite{pfeGithub} \\
    \hline
    \end{tabular}%
  \caption{Configuration and accuracy of each model. A shortened name is provided for later reference. Note that these accuracy values are calculated by our internal verification and may differ slightly from the stated values for each model's source publication (when available). \\
  *Sources not associated with original publication. 
  }
  \label{tab:model-desc}%
\end{table*}%

Ten independently trained face-recognition models of various pedigree and performance were selected for study. 
Models were selected to include a variety of performances, training datasets, CNN architectures, and angular loss functions.

Table~\ref{tab:model-desc} lists the training datasets, architectures, loss functions, performance on IJB-C, and GitHub source for the ten models. 
The training datasets used are described in Table~\ref{tab:dataset-desc}, ranging from 500,000 images in CASIA-WebFace \cite{CASIAwebface} to 17 million images in Glint360K \cite{an2020partial_fc}.
The relative size and complexity of each CNN backbone is listed in Table~\ref{tab:backbones}, ranging from MobileNet's 6 million parameters to ResNet-100's 44.5 million parameters.
For more details on models and datasets, please refer to their respective publications and code sources.

Using four of our worst-performing models, we initially carried out experiments on LFW (see Appendix~\ref{app:lfw}).
While these experiments yielded promising results, performance on LFW is highly-saturated using modern CNNs.
This led us to IJB-C, a more recent public face benchmark dataset which focuses on unconstrained media, and presents a much greater challenge.
IJB-C includes both still images and video frames along with many pre-defined evaluation protocols. 
We focused on 1:1 verification, which includes pairs of multi-image templates and a broad range of difficulties.
Performance on IJB-C is also typically reported as true acceptance rates at fixed false acceptance rates (TAR @ FAR), which allowed us to further quantify relative quality of mappings between networks as expressed through ever greater intolerance for false matches. 

\subsection{Model Evaluation}
\label{ssec:method-eval}
All 10 models were downloaded pre-trained from their respective sources, and evaluated on IJB-C using the 1:1 verification protocol. 
Each model was evaluated using its own source repository's preprocessing steps including face detection and cropping. 
These may differ in crop size, aspect ratio, or similarity transform target, however, the same set of IJB-C images were used in all cases.

Each model was then passed preprocessed images to produce an associated set of embeddings.
Embeddings were all evaluated in the same fashion, using evaluation code adapted from Jia Guo and Jiankang Deng's InsightFace project on GitHub \cite{insightfaceGithub, arcface}.
For all 10 networks here, the dimensionality of the feature space is 512.
However, with the Probabilistic Face Embeddings (PFE) architecture (models M1-64S-PFE and C-64S-PFE) \cite{shi2019probabilistic}, the output of a second uncertainty module was concatenated to features yielding 1024 dimensions.
This uncertainty module consisted of a network with two fully-connected layers with input and output dimensions of 512, equal to the dimension of the output of the base CNN model. 

IJB-C 1:1 verification consists of generating many templates, each corresponding to one or more images.
In cases where video frames are present in a template, features belonging to the same video are first aggregated by simple vector average.
A single template vector was then calculated as an L2-normalized sum of all image and video features within that template.
Template pairs were scored by the inner product (dot product), equivalent to cosine similarity since all templates are unit-length.
Finally, a list of template pairs was used for ROC analysis to determine true acceptance rates at fixed false acceptance rates (TAR @ FAR).
In the case of PFE, this differed from their mean likelihood score, treating $\sigma$ simply as an additional feature, allowing us to maintain a uniform distance measure across all models for later cross-model comparisons.
The performance of each model at 6 FARs is provided in Table~\ref{tab:model-desc}.

\subsection{Calculating Mappings}
\label{ssec:method-map}

We are interested in calculating the extent to which a linear map converts between the features of two networks.
Let $X_E$ and $X_V$ be the 11,856 enrollment and 457,519 verification images belonging to the IJB-C 1:1 Verification protocol.
For a source network $f_A$ and target network $f_B$, we fit a matrix $\matr{M}_{A\to B} \in \mathbb{R}^{d_A \times d_B}$ such that 
\begin{equation}
    f_B(X_E) \approx \matr{M}_{A\to B} f_A(X_E)
\end{equation}
for all input images $x \in \mathbb{R}^{w \times h \times 3}$.
Essentially, this approach seeks a mapping which minimizes the distance between pairs of points in feature space corresponding to the same image, up to differences in preprocessing.
We also explicitly normalized the result of $\matr{M}_{A\to B} f_A(X_E)$ so that it corresponds to the output of the models we studied. 

We calculated mappings using two methods, both using pairs of embeddings generated from the IJB-C 1:1 verification enrollment set.
To elaborate, these 11,856 images were passed to both models to generate 11,856 pairs of embeddings.

\textbf{Linear} mappings were computed by solving the ordinary least squares regression problem over image pairs:
\begin{equation}
    \minimize \sum_{i=1}^m ||\tilde{\matr{M}}_{A\to B} f_A(x) - f_B(x)||_2 .
\end{equation}

\textbf{Rotation} mappings were computed using the methods developed by Wahba and Kabsch for finding the optimal rotation for minimizing the distances between two sets of points \cite{wahba1965least, kabsch1976solution}.
Simply put, this algorithm consists of computing the singular value decomposition of the cross-covariance matrix of two sets of points $f_A(X_E)$ and $f_B(X_E)$, followed by recomposition with all singular values set to 1. 
To ensure no flips or mirroring, the last singular value (corresponding dimension of least variance) is optionally set to -1.
In other terms, we calculated rotation mappings as:
\begin{equation}
\begin{split}
    f_A(X_E)^T f_B(X_E) & = U \Sigma V_h \\
    M_{A\to B} & = U I' V_h
\end{split}
\end{equation}
where $$I' = \mathrm{diag} \left( \left[ 1 \quad 1 \quad \dots \quad 1 \quad \mathrm{det}(U)*\mathrm{det(V_h)} \right] \right).$$
This produced a linear mapping matrix with the additional constraint of being orthogonal and having determinant 1, a rotation.
Note that $f_A(X_E)$ and $f_B(X_E)$ would typically be centered first to find the optimal rotation axes, but we left points in their original translation (on the unit hypersphere), since we intend to rotate about the origin.

\subsection{Evaluating Mappings}
\label{ssec:method-map-eval}
A natural method for mapping evaluation is to measure impact on performance using a validation dataset of faces unseen during training of any models.
Essentially, we produce mapped features from the mapping's source network, and evaluate them against features generated by the mapping's target network.
As in Section~\ref{ssec:method-eval}, templates were generated from collections of embeddings, except each template in a pair was generated by a different network.

To be precise, target model templates were calculated from embeddings in the verification set, $f_B(X_V)$, and source model templates were calculated from mapped embeddings generated from the same images $\matr{M}_{A\to B} f_A(X_V)$.
As in the previous section, template match scores are computed as the inner product, equivalent to cosine similarity when templates are unit-length.
ROC analysis is performed to produce true accept rates at various false accept rates (TAR @ FAR), which may be compared to unmapped model performance.

\section{Cross-CNN Mapping Results}
\label{sec:results}

\begin{figure*}
  \centering
  \renewcommand{\arraystretch}{1.5}
  \includegraphics[width=0.95\textwidth]{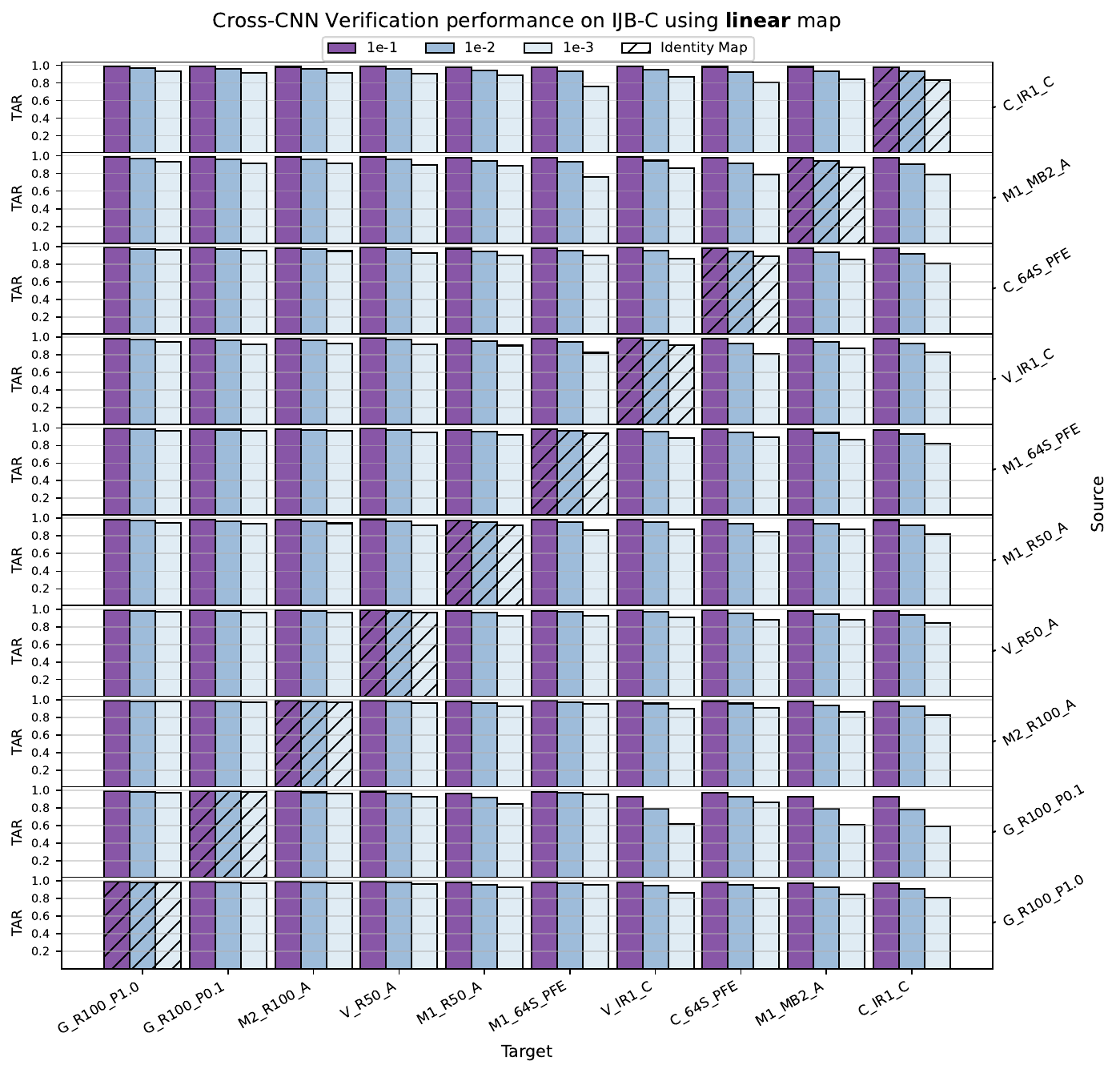}
  \caption{
  \textbf{Linear} maps reveal consistent overlap between feature spaces of distinct CNNs.
   Bars indicate the TAR on IJB-C 1:1 verification at the FAR indicated by the bar color.  Hatched bars correspond to the unmodified performance of each model (also in Table~\ref{tab:model-desc}). Models are sorted according to unmodified TAR at a FAR of 0.01. Off-diagonal bars correspond to the accuracy obtained when comparing features across networks, with the ``Source'' model's features mapped by linear transformation to approximate the ``Target'' model's features. Models are referred to by their ``Short Name'' listed in Table~\ref{tab:model-desc}. Best viewed in color.
  }
  \label{fig:mapping-performance}
\end{figure*}
\begin{figure*}
  \centering
  \renewcommand{\arraystretch}{1.5}
  \includegraphics[width=0.95\textwidth]{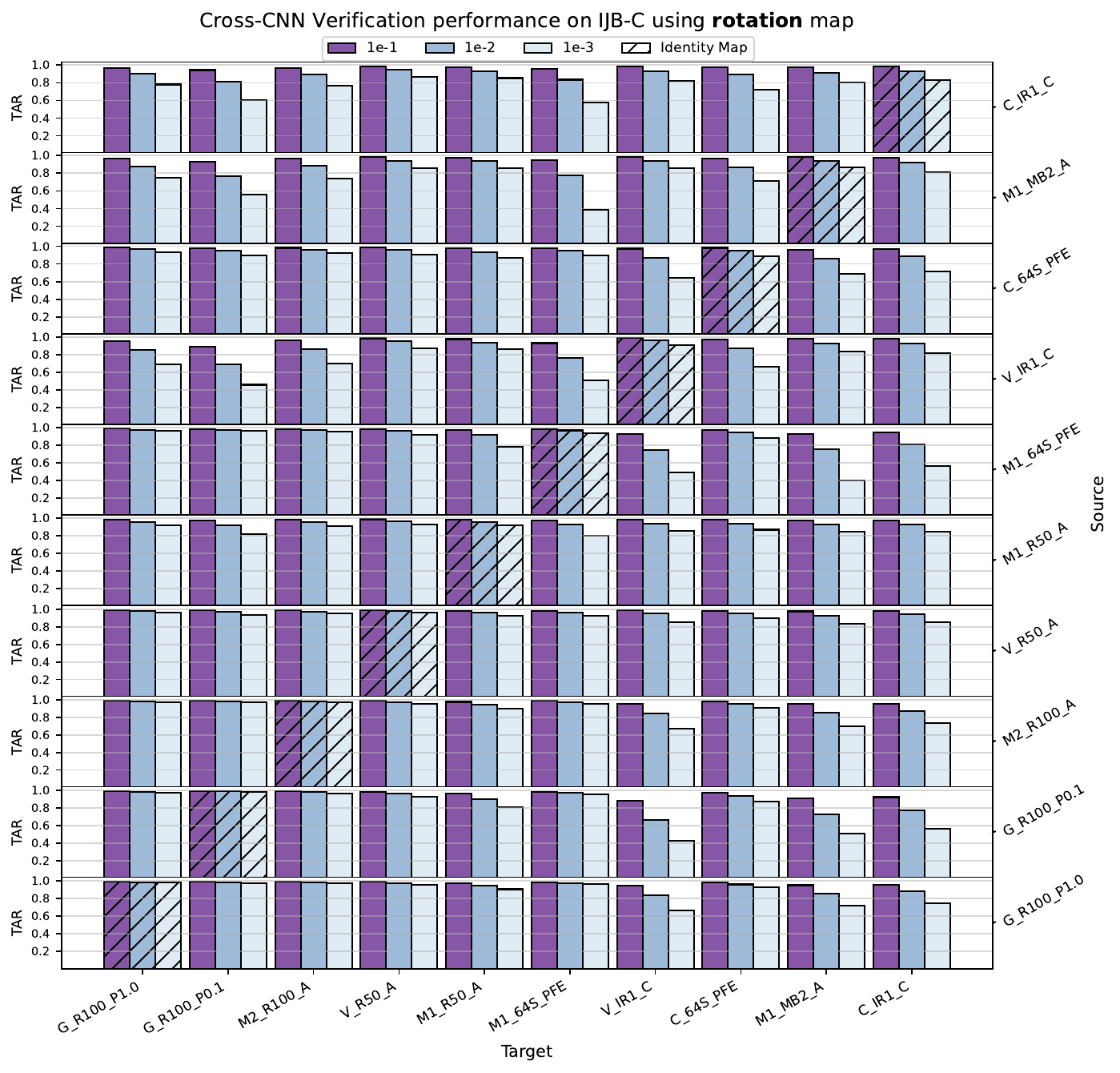}
  \caption{
  \textbf{Rotation} maps also reveal consistent overlap between feature spaces of distinct CNNs. See the caption of Figure~\ref{fig:mapping-performance} for figure description.
  }
  \label{fig:mapping-performance-rotation}
\end{figure*}

Mapping evaluation results are summarized in Figure~\ref{fig:mapping-performance}.
Hatched bars along the diagonal show the same TARs listed in Table~\ref{tab:model-desc}.
Off-diagonal elements contain TARs produced by cross-CNN evaluation as described in Section~\ref{ssec:method-map-eval}, with the row label indicating the source model, and the column label indicating the target model.
These labels correspond to each model's ``Short Name'' in Table~\ref{tab:model-desc}.
Rotation mapping evaluations are in Figure~\ref{fig:mapping-performance-rotation}.

For the bulk of cross-CNN comparisons, linear mappings seem to convert embeddings effectively and with little performance penalty.
When looking for poor mapping performance, Partial FC loss using 10\% label subsampling (G\_R100\_P0.1) seems to produce features which are more difficult or dissimilar.
In contrast, the same model trained using all of Glint360K's labels (G\_R100\_P1.0) consistently produced high performance when mapped (though not necessarily the highest).
This suggests that label subsampling, while impacting single-model verification performance very little, has a relatively large effect on embedding space similarity.
The contrast in mapping performance between these two models which themselves are highly compatible and near identical in training setting prompts further investigation, as explained in Section~\ref{sec:discussion}.

Even when constrained to only rotate embedding spaces, cross-CNN performance is still at or near single-CNN performance at a FAR of 0.1.
While this FAR is very weak, these results provide a demonstration that mapping between face verification CNNs is possible using linear or rotation maps.
Interestingly, some mappings exceed the performance of their target or source model (but not both).
Compared with linear maps, rotation produce a higher penalty in almost all cases, and the penalty was nearly symmetric when source and target are swapped, due to the structure of finding an orthonormal solution.
The extra constraint of rotation maps further reveals the nature of cross-CNN relationships.

As the FAR is decreased to 1e-3, we maintain high TARs for the most successful networks, but networks with less representational complexity have less success in representing the more complex networks. 
Smaller FARs are listed with our complete results in Figures~\ref{fig:mapping-performance-all-linear} and \ref{fig:mapping-performance-all-rotation}, including some failure cases which indicated an exciting direction for future work, as discussed in Section~\ref{sec:discussion}.

\subsection{Sensitivity to number of images}
\label{ssec:sensitivity}
\begin{figure*}
  \centering
  \includegraphics[width=1.0\textwidth]{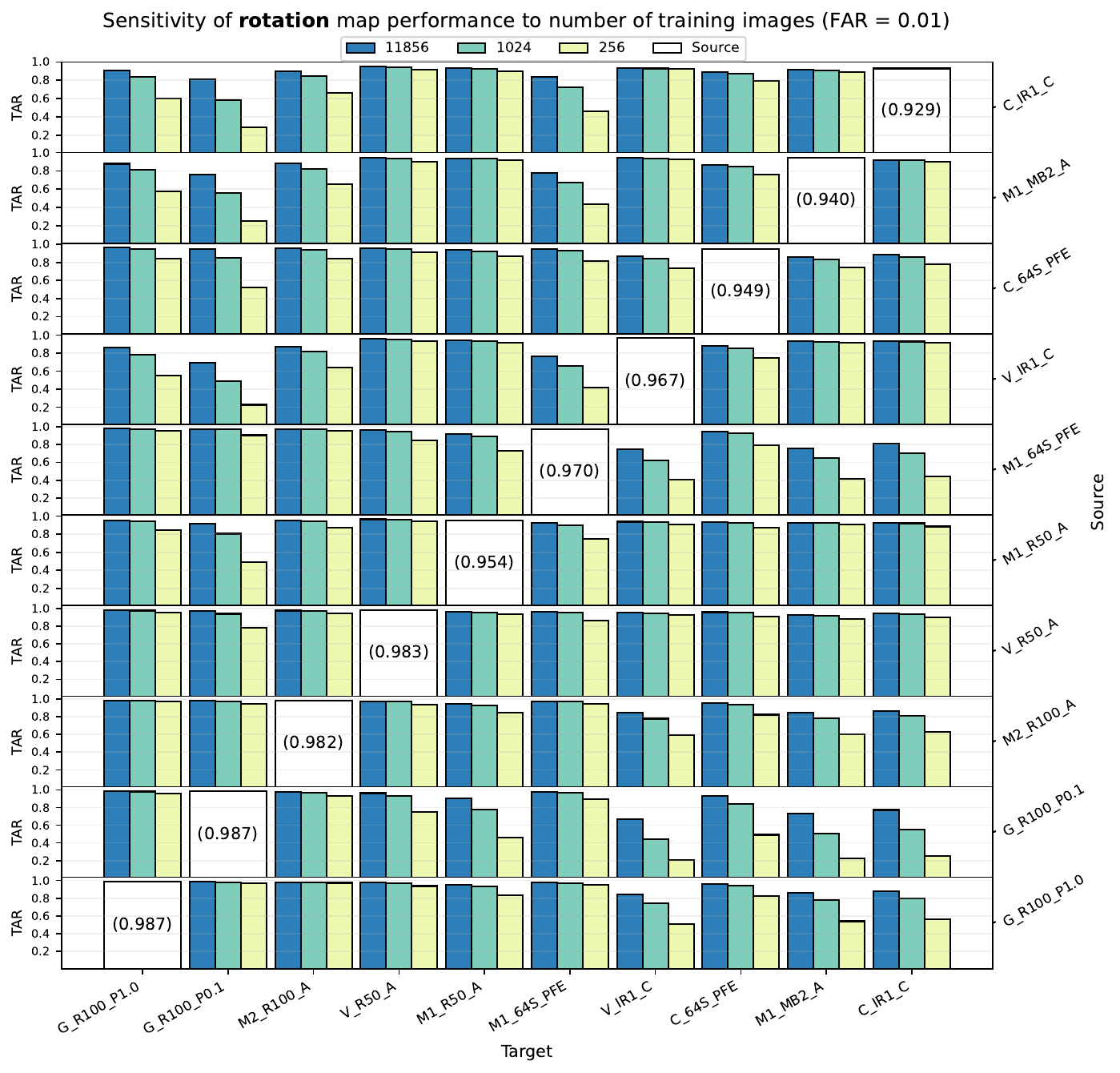}
  \caption{ 
  Each bar represents the TAR at a FAR of 0.01 achieved by comparing "Source" CNN features to "Target" CNN features after rotation mapping. TAR is shown for rotation mappings computed from 256, 1024 and 11,856 pairs of corresponding embeddings.
  }
  \label{fig:mapping-sensitivity}
\end{figure*}

To better understand the complexity of mappings between embedding spaces, we also fit and evaluate mappings using a variable number of paired examples.
Specifically, we selected a random subset of the 11,856 images in the IJB-C 1:1 verification enrollment set, and fit mappings as described before, using a smaller set of images selected at random.
Then, mappings were evaluated as before, by comparing the mapped embeddings of one network to the unmapped embeddings of another on the IJB-C 1:1 verification set, after both sets of embeddings have been aggregated into templates.
Average performance was collected over 3 repetitions of this experiment using independent random subsets.
The resulting TARs at a FAR of 0.01 using all (11,856), 1024, or 256 samples to fit said mappings is illustrated in Figure~\ref{fig:mapping-sensitivity}.
Rotation mappings were used, as they have fewer degrees-of-freedom and thus require fewer examples.
For more sample sizes, see Figure~\ref{fig:mapping-sensitivity-all} in Appendix~\ref{sec:full-figures}.

Though some mappings performed worse with fewer examples, many provided near unchanged performance.
As before, the relative difference in model pairs seem to generally reduce mapping performance.
Since Figure~\ref{fig:mapping-sensitivity} is sorted by single-model performance model pairs closer to the top-left and bottom-right of this figure are further from each other in relative performance, and also seem to produce generally poorer performance when mapped.

These results are broadly encouraging, further confirming the presence of fundamental similarity between embedding spaces. 
On the other hand, cases where performance is heavily impacted when using fewer examples reveal the relative difficulty of certain CNN pairs over others.

\section{Security Implications}
\label{sec:security}
Anyone handling embeddings from an operational face recognition system based upon existing neural networks must ask themselves this question:

\begin{quote}
    What risks might ensue if embeddings from my system become available to others?
\end{quote}
\noindent
In light of what we have just presented, risks include naming an individual associated with an "anonymous" embedding and possibly even enabling an impersonation attack. 

To recap and setup for explaining these risks, the experiments above demonstrated that two face embedding systems (Systems A and B) are likely to produce the same embeddings, differing only by a fixed linear transformation. Further, these linear transformations can be determined using relatively few (hundreds) paired embeddings. In other words, if embeddings from System A can be obtained, and their corresponding embeddings in System B discovered (e.g. by also obtaining the faces or identities they represent), then the general linear mapping between embeddings spaces can be calculated. Note also that to establish a mapping between systems, the paired embeddings themselves do not actually need to be labeled. To be clear, they must be of the same person, but the actual identity of the person is not itself used. 

Now assuming the mapping between embedding spaces has been determined, how might it be used?  Let us highlight two examples. The first is the risk that, even though those responsible for System A might hope an unlabeled embedding is anonymous, there is a clear path to discovering the withheld identity.  If System B is tied to a database of labeled faces containing the identity of the ``anonymous'' embedding from System A, then a simple identity query to System B with the mapped embedding will return the associated identity. 

For the second example,  let us begin with the ``Template Reconstruction Attack'' described in Mai et al \cite{mai2018reconstruction}.
In this scenario, an attacker gains access to a face verification model (System A) and embeddings corresponding to a target individual.
Then, the attacker trains a face reconstruction model using System A which converts embeddings into 3-D face representations.
Once embeddings can be reconstructed into faces, the attacker can reconstruct the face of a target individual and impersonate them when authenticating using the compromised system (System A).
Why the ``Template Reconstruction Attack'' approach is important here is because, when combined with our work, direct access to System A may no longer be necessary. Instead, a single embedding from System A can be mapped into System B's embedding space and then the reconstruction process proceeds using System B.

Our first example is an approach actually put into practice by us in the context of processing face embeddings from two major participants in the DARPA AIDA program. Each of these participants generated embeddings from a set of images and video frames, associating embeddings with random unique identifiers. We were given these embeddings alongside other document extractions in a non-reversible fashion to simulate conditions where primary sources need remain anonymous (e.g. whistleblowers). Our goal was to combine such multi-source extractions into a knowledge base for downstream participants. By pairing embeddings from different AIDA participants with the same unique identifier (thus the same source image and face), we were able to fit linear mappings between the embeddings of unknown models. Ultimately, this allowed us to create correspondences between faces which we had never seen, embedded by models we could not access. The second example, the possibility for enabling an impersonation attack, is not something we have tried and we consider doing so beyond the scope of this paper. That said, the idea is well founded and clearly worthy of further exploration. 

In closing, we note that these vulnerabilities depend upon the reliability of the mapping.
As we demonstrate in Figure~\ref{fig:mapping-sensitivity}, some mappings perform poorly, especially when using fewer examples. 
Still, from the perspective of security, even a low-reliability vulnerability is worth consideration.
More to the point, our goal in this paper has not been to present detailed end-to-end attack models. Instead, we are emphasizing that embeddings --- despite being the result of extremely non-linear upstream processes --- are not alone sufficient to hide identity.
In summary, face embeddings and their associated identities should be treated as sensitive information and stored as such.
Securing face embeddings against the vulnerability we describe here could be accomplished by an invertible, sufficiently nonlinear transformation applied before storage. 
For example, strong encryption of embedding vectors should be adequate to guard against the risks cited above.

\section{Discussion}
\label{sec:discussion}
We are confident that the results presented here are strong evidence for a fundamental similarity between common CNN-based face recognition systems.
Though we only study 10 models, these results suggest that the similarity structure of face CNN embeddings remains stable despite changes in architecture, dataset, and loss.
Although some information does not transfer in these linear or rotation maps, clearly the bulk of information does.
Where one might expect a new combination and configuration of CNN layers and training examples to produce new learned representations, our work demonstrates that embeddings produced by CNN models trained for the same task approximate a \textbf{canonical} space, which each model appears to converge upon.

Generally, the performance of the poorest-performing model in a pair seems to provide an upper bound of cross-CNN performance.
However, certain cases perform worse, suggesting certain models have greater compatibility than others, or even that some information is not captured by poorer-performing models.
For example, take G\_R100\_P0.1 (ResNet100 trained on Glint360k using Partial FC loss) which performs comparatively well on its own, and in many cross-CNN mappings.
This model is already somewhat distinct in that it uses a random subset of 10\% of identities for computing softmax loss during training \cite{an2020partial_fc}.
Take note, however, of the target models which produce poor mapping performance, V\_IR1\_C, C\_IR1\_C, and M1\_MB2\_A. 
While these models perform worse already, they seem to have greater incompatibility with G\_R100\_P0.1 than others.
Roughly speaking, if model A is compatible with model B, and model B is compatible with model C, then why isn't A compatible with C?
We hypothesize that, despite evidence for broad task-driven similarity between CNNs, certain design features of these CNNs do impact finer aspects of their representation. 
In other words, it may be that subsampling identities during training as with Partial FC biases the resulting model away from a representation found using InceptionResNet or MobileNet architectures. 

While we demonstrate results using 90 pairs of CNNs, it's unclear how small variations in preprocessing or model implementation may contribute to mapping performance.
While we include many architectures and results for a large face dataset, 
finer characterization of these relationships requires carefully controlling for architecture and training details.
Such analysis will likely come at great resource cost, so we leave a systematic exploration with increased granularity and controls to future works.

Further, our method may be explicitly lossy when features are represented with different numbers of dimensions, producing non-square mapping matrices (i.e. non-zero nullity per the rank--nullity theorem). 
The work of Gong~\etal\cite{gong2019intrinsic} offers insight along these lines by providing evidence for a far reduced "intrinsic dimensionality" produced by common face verification CNNs.
This suggests that while rectangular linear transformations indeed project information into fewer dimensions, mapping between two sets of face embeddings may require far fewer dimensions than the maximum rank of a rectangular matrix.

\section{Conclusion}
\label{sec:conclusion}


The existence of performance-preserving linear mappings between face recognition CNNs which vary in training and construction suggests this phenomenon depends primarily--if not solely--upon the modeling task. 
Subsequently, the existence of task-dependent canonical embedding spaces suggests there is a high degree of redundancy in the training procedure of bespoke CNN-based models for a common task, as the bulk of the learned representation is unchanged.  
Consider the computational effort required to train and develop these models, given that they each converge to highly similar solutions. 
Perhaps there is a richer training signal to-be-developed which takes advantage of this seemingly universal similarity structure, efficiently guiding training towards solutions found by previous models.  
Regardless, these results seem to discourage efforts to optimize CNN architectures for the purpose of improved performance, so long as the modeling task (i.e. the dataset) provides the biggest impact on the content of modeling output.

More broadly, if the modeling task most significantly impacts the structure of embedding space, it prompts further investigations into the relationship between these spaces and the modeling tasks which produced them.
Others have investigated the relationships between modeling tasks, finding where learned representations of one task help or hurt performance on another task \cite{zamir2018taskonomy, zamir2020robust}.
Besides the impact on performance, cross-task representations could also be studied by their interactions in embedding space structure.
Clearly, more work is necessary to understand how these structures are organized, as our own work depends upon performance as a downstream measure of similarity.
Still, if we want models to learn fundamentally new representations, our work suggests they need new tasks.

\appendices



\section{Pilot experiment on LFW}
\label{app:lfw}
\begin{figure}
    \centering
    \includegraphics[width=0.48\textwidth]{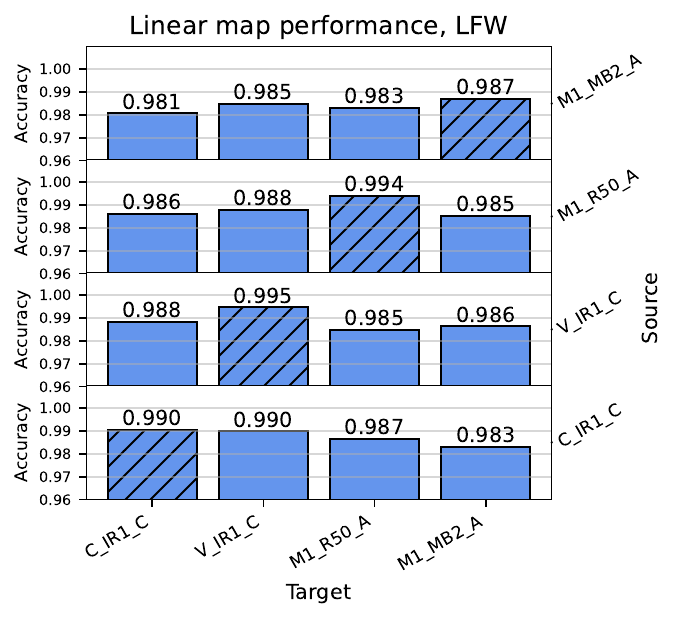}    
    \caption{Linear mappings are sufficient for converting between the features of different CNNs with minimal drop in performance.
    Cross-hatched bars indicate the performance when evaluating a model against its own embeddings, without mapping.}
    \label{fig:lfw}
\end{figure}
\begin{figure}
    \centering
    \includegraphics[width=0.48\textwidth]{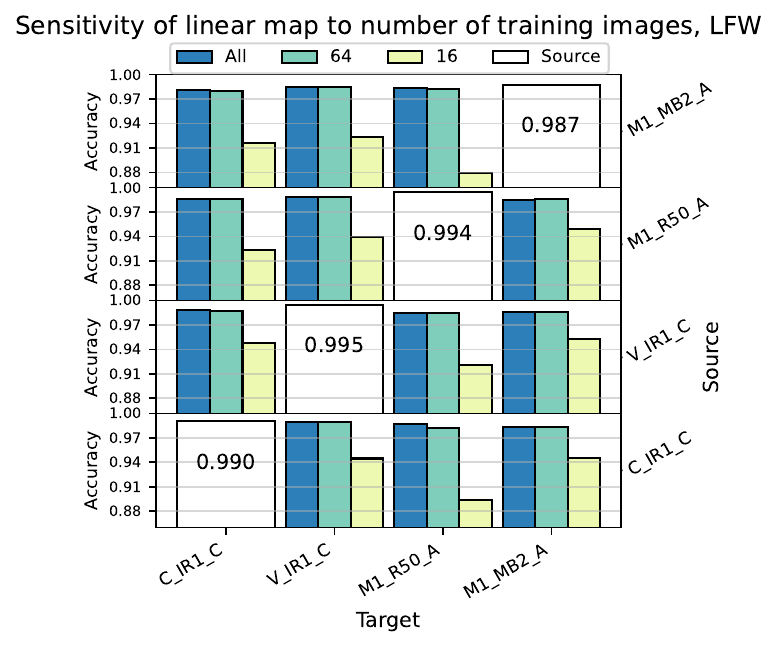}
    \caption{When using a reduced number of images, linear mappings still convert between the features of different CNNs with high performance.}
    \label{fig:lfw-sensitivity}
\end{figure}

LFW verification consists of generating pairs of embeddings for 6,000 predefined image pairs.
Then, accuracy is measured by performing 10-fold cross-validation as specified by the \textit{Unrestricted, labeled outside data} protocol documented in LFW \cite{labeledfacesinthewild}.
Accuracy on LFW is highly-saturated by modern CNNs, so we evaluate four of our worse-performing models. 
For this performance, see the hatched bars of Figure~\ref{fig:lfw}.

Linear mappings are fit using the same method described in Section~\ref{ssec:method-map}, and evaluated using the same method described in Section~\ref{ssec:method-map-eval}, instead reporting the mean accuracy obtained on each of the ten cross-validation folds.
More precisely, one fold is held out for testing while the remaining nine are used to find a linear mapping.
Then, the held out fold is used for cross-CNN evaluation.
This process is repeated ten times, once for each fold.
The mean accuracy obtained on held out folds is reported in Figure~\ref{fig:lfw}.

Sensitivity experiments can also be carried out as in Section~\ref{ssec:sensitivity}, using a random subset of embedding pairs. 
For each fold, a mapping is fit on a random subset of training fold embeddings, and evaluated on the remaining test fold.
The mean accuracy obtained at different sizes of training subsets is reported in Figure~\ref{fig:lfw-sensitivity}.

Though these experiments are carried out on an easier dataset, they suggest that this linear correspondence phenomenon is not necessarily dependent on evaluation method or dataset.

\section{Full Figures}
\label{sec:full-figures}
\begin{figure*}
    \centering
    \includegraphics[width=1.0\textwidth]{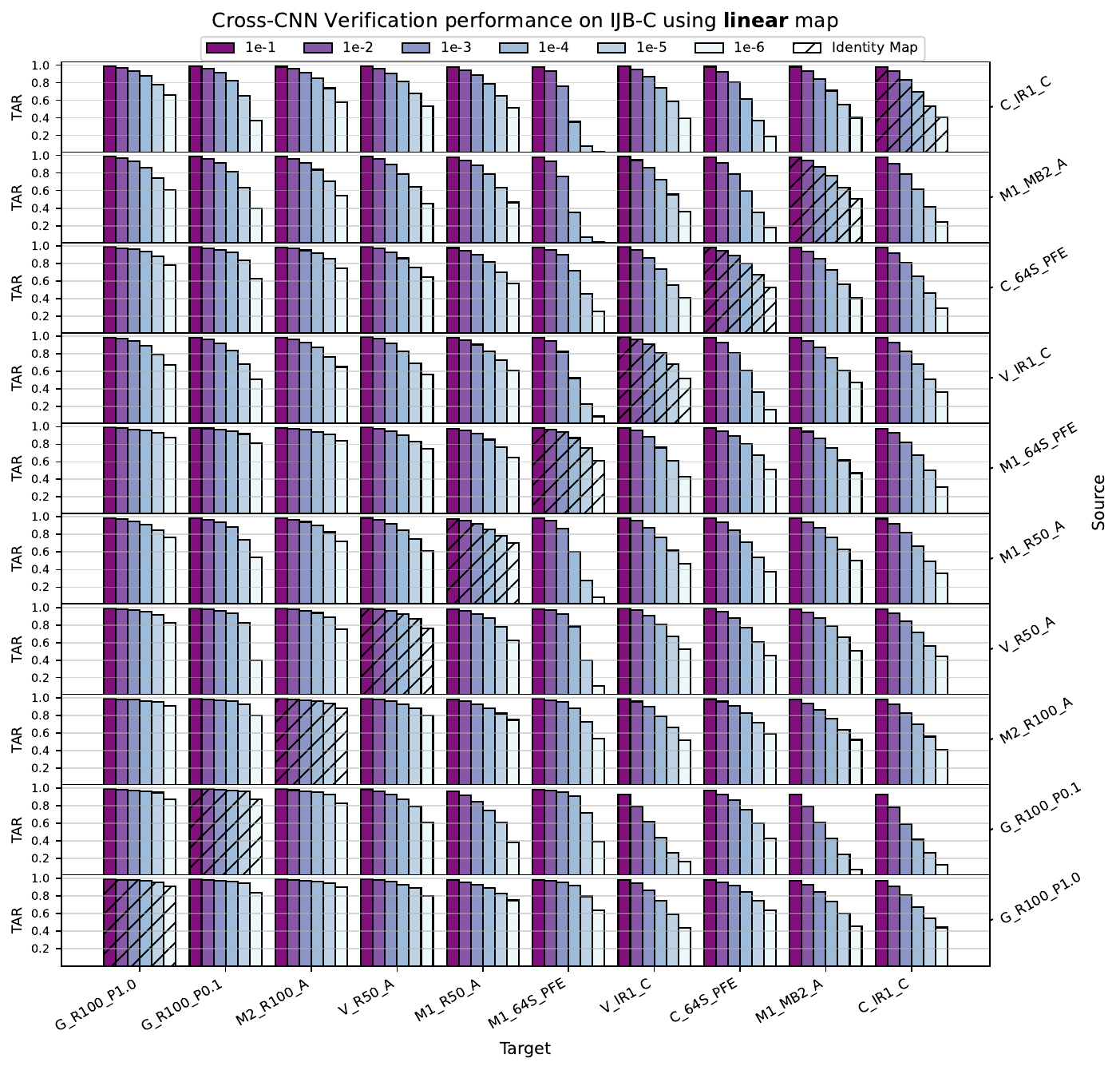}
    \caption{Full \textbf{linear} mapping performance results, in the same format as Figure~\ref{fig:mapping-performance}.}
    \label{fig:mapping-performance-all-linear}
\end{figure*}

\begin{figure*}
    \centering
    \includegraphics[width=1.0\textwidth]{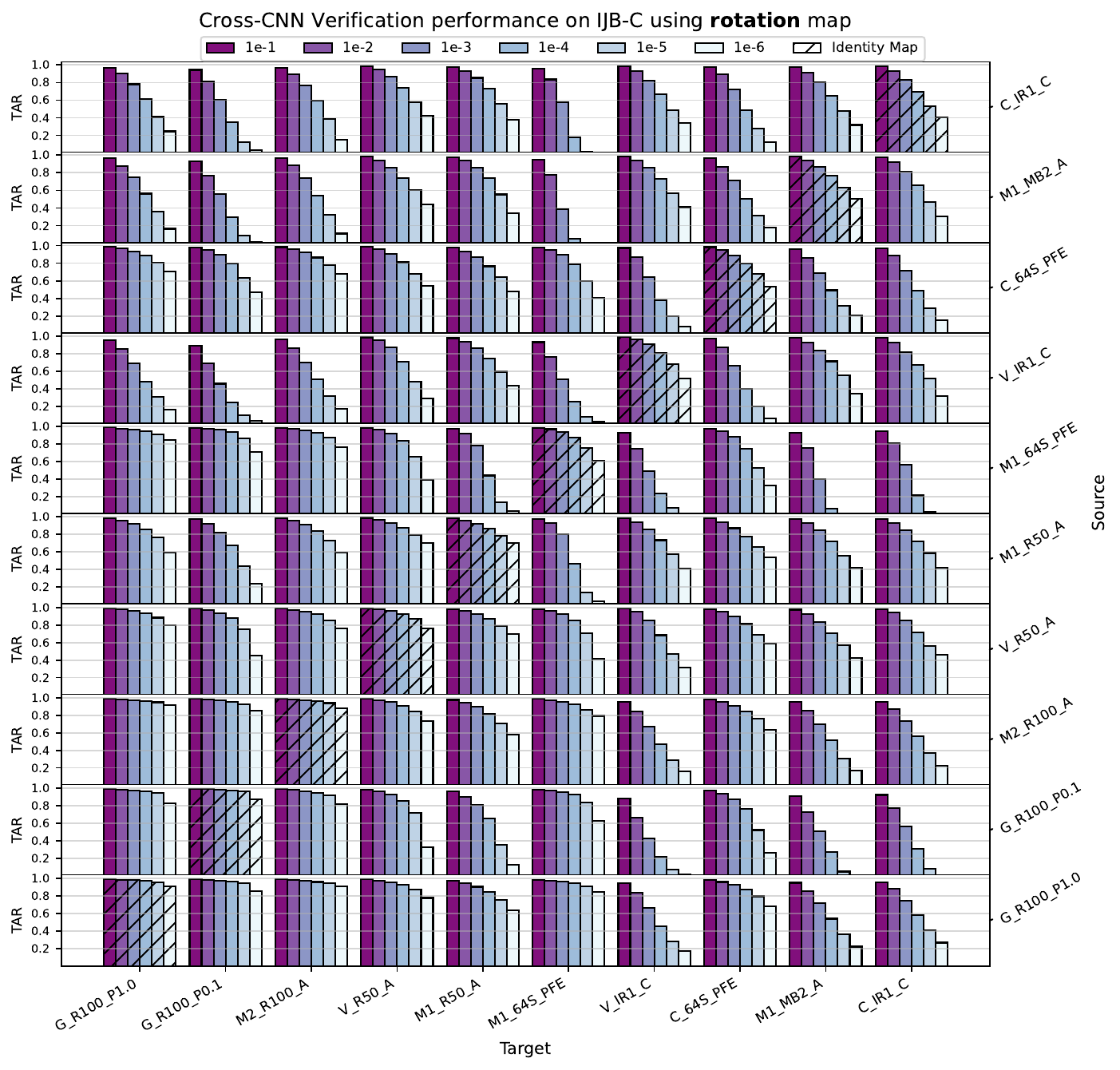}
    \caption{Full \textbf{rotation} mapping performance results, in the same format as Figure~\ref{fig:mapping-performance-rotation}}
    \label{fig:mapping-performance-all-rotation}
\end{figure*}

\begin{figure*}
  \centering
  \renewcommand{\arraystretch}{1.5}
  \includegraphics[width=0.8\textwidth]{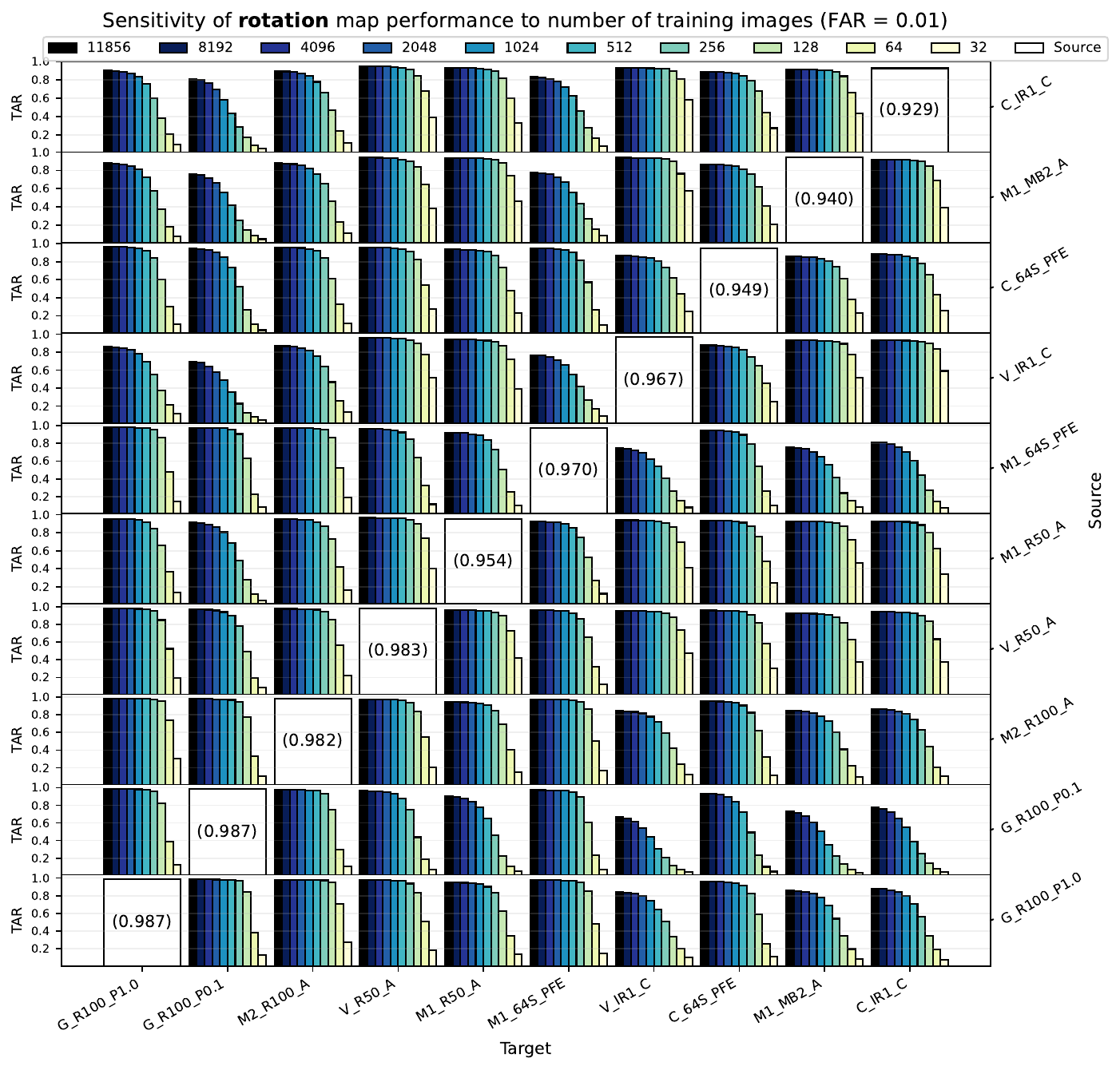}
  \caption{
  Full mapping sensitivity results, including the same data presented in Figure~\ref{fig:mapping-sensitivity} and following the same format.
  All performances represented here are generated from mapped features (i.e. no single-model performance is included).
  }
  \label{fig:mapping-sensitivity-all}
\end{figure*}

We provide figures containing the full results from our cross-CNN IJB-C 1:1 verification experiments using a linear map (Figure~\ref{fig:mapping-performance-all-linear}), a rotation map (Figure~\ref{fig:mapping-performance-all-rotation}), and a restricted training sample size using a rotation map (Figure~\ref{fig:mapping-sensitivity-all}). 


\ifCLASSOPTIONcompsoc
  \section*{Acknowledgments}
\else
  \section*{Acknowledgment}
\fi

This work was supported by a subaward from DARPA FA8750-18-2-0016-AIDA – RAMFIS: Representations of vectors and Abstract Meanings for Information Synthesis. 
This work does not express the opinions of the sponsor.
Thank you to Michelle Wern for designing Figure~\ref{fig:teaser}.

\ifCLASSOPTIONcaptionsoff
  \newpage
\fi


\bibliographystyle{IEEEtran}
\bibliography{IEEEabrv,MAIN}

\begin{IEEEbiography}[{\includegraphics[width=1in,height=1.25in,clip,keepaspectratio]{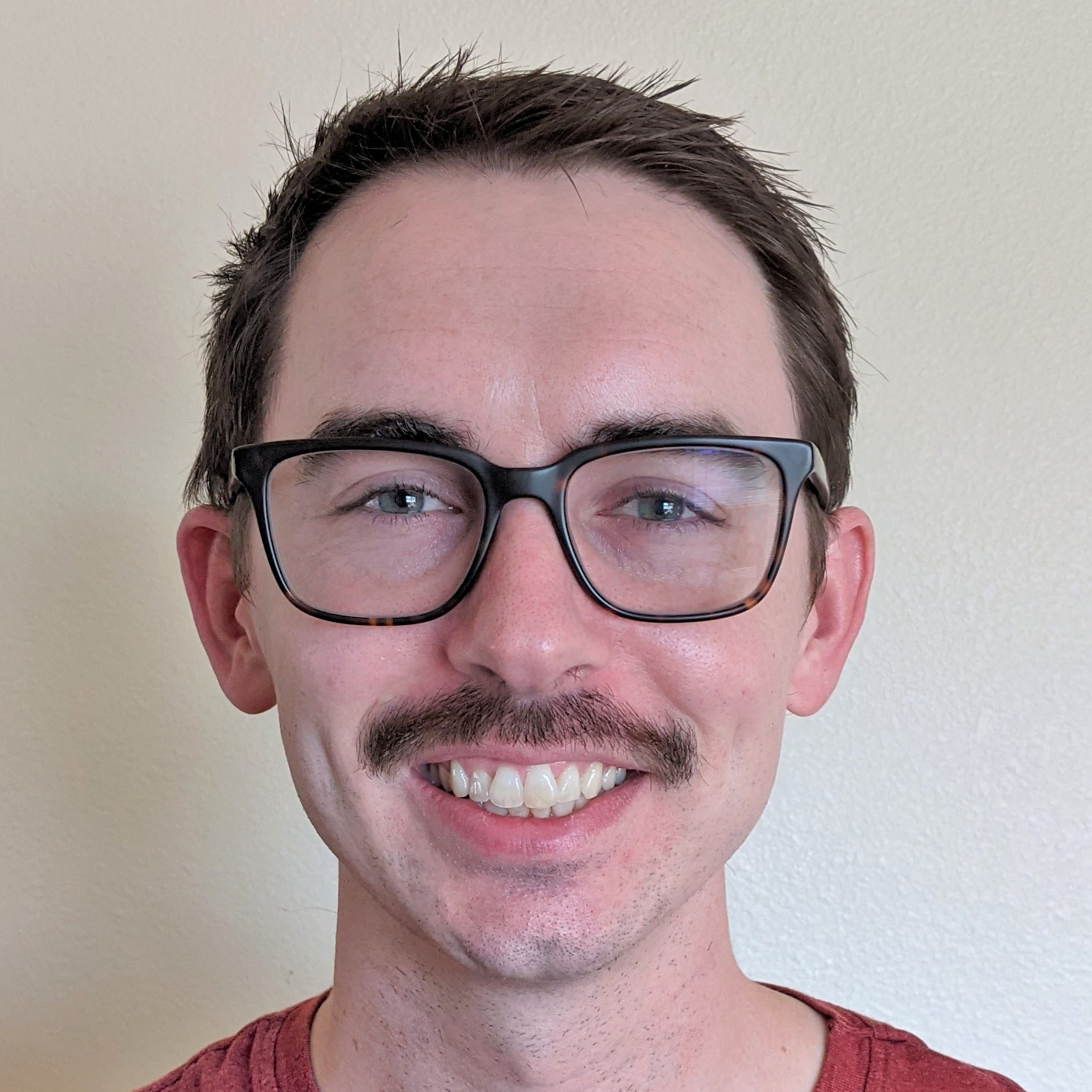}}]{David McNeely-White}
David received a B. Arts degree in computer science from Maryville College, Tennessee in 2014, and a M.Sc. degree in computer science from Colorado State University in 2020, where he is currently pursuing a Ph.D. with the Computer Vision Group. 
His research interests include deep neural network representational similarity, and realtime human perception for virtual agents (gesture, pose, affect, and speech).
\end{IEEEbiography}

\begin{IEEEbiography}[{\includegraphics[width=1in,height=1.25in,clip,keepaspectratio]{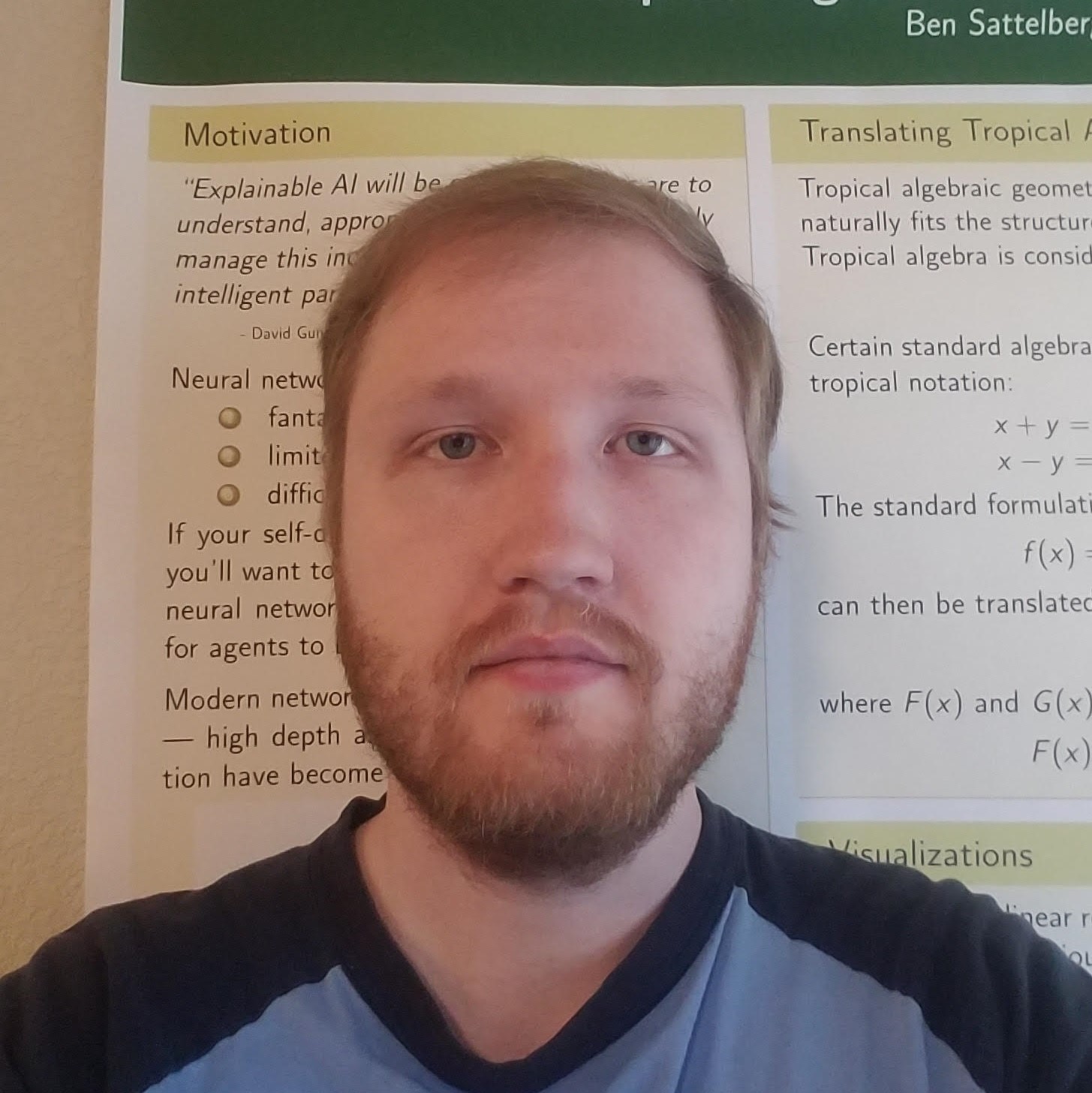}}]{Ben Sattelberg}
Ben received Bachelor and Master of Science degrees in Applied and Computational Mathematics from Colorado School of Mines in 2015 and 2016.  He is currently pursuing a Ph.D. at Colorado State University with the Computer Vision Group.  His research interests include neural network explainability and visualization. 
\end{IEEEbiography}

\begin{IEEEbiography}[{\includegraphics[width=1in,height=1.25in,clip,keepaspectratio]{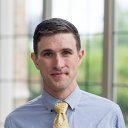}}]{Nathaniel Blanchard}
Dr. Blanchard received his Ph.D. in Computer Science and Engineering from the University of Notre Dame in 2019. The same year, he joined the Computer Science department at Colorado State University as an Assistant Professor. He has published 21 papers; some of his accolades include publishing at the top computer vision conference (CVPR) and best student paper at WACV 2021. 
\end{IEEEbiography}

\begin{IEEEbiography}[{\includegraphics[width=1in,height=1.25in,clip,keepaspectratio]{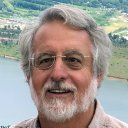}}]{Ross Beveridge}
Dr. Beveridge received his Ph.D. in Computer Science from the University of Massachussetts in 1993. He is presently a Professor in the Department of Computer Science at Colorado State University working broadly in the area of Computer Vision. He has published over 150 refereed papers which have been cited over 9,000 times. Professor Beveridge, working closely with Jonathon Phillips at NIST, pioneered new approaches to characterizing how human face recognition algorithm performance varies with demographics attributes such as age, race and gender. Professor Beveridge's current work is motivated by the promise of visually aware agents. Working jointly with colleagues at Brandeis University he has overseen the development of an AI agent able to interact with a person using both speech and sight.
\end{IEEEbiography}

\end{document}